# Multimodal Data Integration for Sustainable Indoor Gardening: Tracking Anyplant with Time Series Foundation Model

Seyed Hamidreza Nabaei, M.S.,[1] Zeyang Zheng, B.E.,[2] Dong Chen, Ph.D.[3]
[4]Arsalan Heydarian, Ph.D.[3]

[1]Department of Systems and Information Engineering, Link Lab, University of Virginia, Charlottesville, VA 22903; e-mail: fgx9eq@virginia.edu
[2]Department of Computer Engineering, Link Lab, University of Virginia, Charlottesville, VA 22903; e-mail: yuq8cp@virginia.edu
[3]Agricultural and Biological Engineering, Mississippi State University, Starkville, MS, 39762, USA; e-mail: dchen@abe.msstate.edu
[4]Department of Civil and Environmental Engineering, Link Lab, University of Virginia, Charlottesville, VA 22903; e-mail: ah6rx@virginia.edu

**ABSTRACT**

Indoor gardening within sustainable buildings offers a transformative solution to urban food security and environmental sustainability. By 2030, urban farming, including Controlled Environment Agriculture (CEA) and vertical farming, is expected to grow at a compound annual growth rate (CAGR) of 13.2% from 2024 to 2030 according to market reports. This growth is fueled by advancements in Internet of Things (IoT) technologies, sustainable innovations such as smart growing systems, and the rising interest in green interior design. This paper presents a novel framework that integrates computer vision, machine learning (ML), and environmental sensing for automated monitoring of plant health and growth. Unlike previous approaches, this framework combines RGB imagery, plant phenotyping data, and environmental factors such as temperature and humidity to predict plant water stress in a controlled growth environment. The system utilizes high-resolution cameras to extract phenotypic features, such as RGB, plant area, height, and width while employing the Lag-Llama time series model to analyze and predict water stress. Experimental results demonstrate that integrating RGB, size ratios, and environmental data significantly enhances predictive accuracy, with the Fine-tuned model achieving the lowest errors (MSE = 0.420777, MAE = 0.595428) and reduced uncertainty. These findings highlight the potential of multimodal data and intelligent systems to automate plant care, optimize resource consumption, and align indoor gardening with sustainable building management practices, paving the way for resilient, green urban spaces.

## INTRODUCTION

The increasing global population, rapid urbanization, and climate change have amplified the demand for innovative and sustainable food production systems. As traditional agriculture faces challenges related to land availability, resource efficiency, and environmental degradation, alternative solutions like Controlled Environment Agriculture (CEA) and vertical farming have emerged as viable options for urban food production. Recent projections estimate that the global indoor farming market size is expected to grow at a compound annual growth rate (CAGR) of 13.2% from 2024 to 2030 while producing 100–180 million tonnes of food annually, providing substantial ecosystem services and energy savings, signaling a significant shift toward indoor agriculture in urban spaces. These systems integrate seamlessly with urban infrastructure, addressing space constraints while ensuring food security and minimizing environmental impact(Das et al., 2022, Clinton et al. 2018, Global Indoor Farming Market Report 2022).

Key technological advancements have accelerated the adoption of indoor farming systems. The integration of Internet of Things (IoT) technologies, environmental sensors, and artificial lighting systems has revolutionized plant monitoring and care. IoT-based environmental monitoring systems allow precise regulation of critical factors such as temperature, humidity, and air quality, which directly influence plant growth and health (Popescu et al. 2024, Kamruzzaman et al. 2019). For example, studies have demonstrated the potential of IoT systems to enhance environmental control in smart indoor gardens while reducing resource waste (Ullo et al. 2020, Udutalapally et al. 2022). These technologies enable real-time monitoring and management, paving the way for sustainable indoor agriculture.

Simultaneously, advancements in artificial intelligence (AI), particularly machine learning (ML) and computer vision, have unlocked new possibilities for automated plant health monitoring. Computer vision techniques such as image segmentation, object tracking, and phenotypic analysis can extract key visual traits, including leaf area, color (RGB values), and growth rate, providing valuable insights into plant health (Kirillov et al.2023, Ravi et al. 2024). For instance, when green plants experience water stress, chlorophyll levels decrease (Yuan et al. 2016). As chlorophyll significantly contributes to the reflection of green light, this reduction leads to a corresponding decline in the green channel value, providing a measurable indicator of plant stress.

Combined with ML models, these methods can deliver predictive analytics to anticipate plant care needs, such as irrigation scheduling and nutrient management. Recent studies have applied machine learning algorithms, such as convolutional neural networks (CNNs) and LSTM, to achieve high accuracy in plant disease detection and care optimization (Pawar et al. 2024, Liu et al. 2024). For instance, a study utilizing IoT and computer vision achieved remarkable accuracy in predicting plant watering needs under controlled environments (Rajendiran et al. 2024, Oudah et al. 2024).

While indoor sensors track environmental conditions like temperature, humidity, and VOC levels, they often function independently of phenotypic data such as plant size and RGB imagery. This separation limits the ability to predict plant health by ignoring the interaction between environmental and phenotypic factors. Bridging this gap by combining environmental sensing with plant phenotyping and predictive analytics is essential to optimize resource efficiency and automate plant care processes. This paper addresses the question: How can integrating environmental and phenotypic data improve the accuracy and scalability of predictive models for plant health monitoring in indoor environments?

To address this challenge, this paper proposes an integrated framework that combines environmental sensing, computer vision, and ML-based predictive analytics for automated plant health and growth monitoring. The system employs commercial environmental sensors to monitor indoor air quality (IAQ)—including VOCs, $CO_2$, temperature, and humidity— while a high-resolution camera captures plant traits such as RGB values. Computer vision techniques, such as image segmentation and object tracking, are utilized to extract phenotypic features, such as plant Height, width, and area, which are further analyzed using Lag-Llama foundation time series models (Rasul et al., 2024). The combination of environmental and phenotypic data enables precise prediction of plant care needs, such as irrigation scheduling and environmental adjustments, with high accuracy.

Preliminary experiments conducted on basil plants in controlled environments demonstrate the effectiveness of the proposed framework. By leveraging multimodal data, including RGB imagery, physiological ratios, and environmental factors, the system significantly improves the accuracy and reliability of predicting water stress for monitoring plant health conditions. Fine-tuned models achieved the lowest errors (MSE = 0.420777, MAE = 0.595428), showcasing the potential of this approach to optimize resource consumption and automate plant care. Integrating such advanced monitoring systems into sustainable building management practices offers a transformative pathway to enhancing urban food security and fostering resilient, green infrastructure.

This paper contributes to the growing body of research on intelligent systems for urban agriculture and offers a comprehensive solution to modern challenges in indoor gardening. By bridging the gap between environmental sensing and AI-driven phenotyping, the proposed framework positions indoor agriculture as a cornerstone of sustainable and energy-efficient urban infrastructure]. The findings underscore the role of integrated AI and IoT systems in fostering smarter, greener, and more resilient urban spaces, addressing critical challenges in global food security and environmental sustainability.

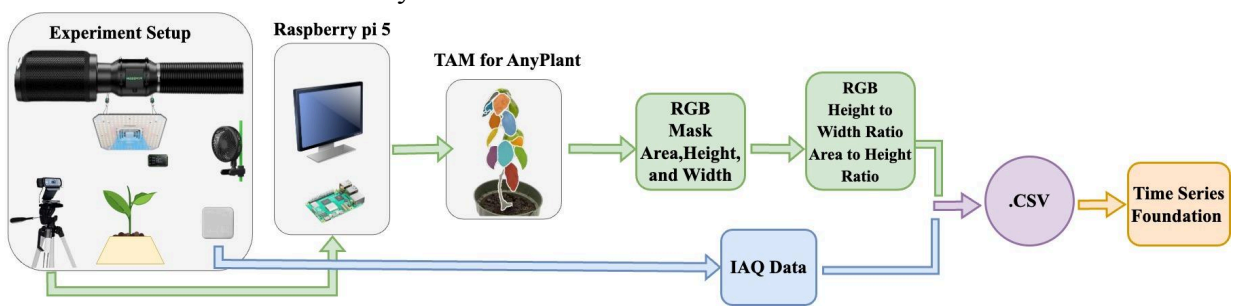

Figure 1. Workflow of the Experimental Framework for Automated Plant Monitoring

**METHODOLOGY**

**Experimental Setup** The experiments were conducted using a Vivosun Smart Grow Tent Kit, which provided a controlled indoor environment to monitor the health and growth of a basil plant. The grow tent was equipped with an LED grow light adjusted to an intensity of 50%.

**Data Acquisition**
1. **Environmental Data Collection** Environmental parameters were continuously monitored using an IAQ Omni Awair Sensor placed within the tent. This sensor measured temperature, humidity, light intensity, VOC (Volatile Organic Compounds), $CO_2$ levels, and PM2.5 particulate matter at regular intervals. These data provided insights into the conditions surrounding the basil plant and were essential for correlating environmental stressors with plant health.
2. **Visual Data Collection** To capture high-resolution images of the basil plant, a common 1080 web camera was positioned in front of the plant to provide a fixed front-view perspective. The camera was connected to a Raspberry Pi 5, which was programmed to take images automatically at 10-second intervals. This process continued for 1.5 months, generating a comprehensive dataset of visual observations. Randomized watering schedules were applied during the experiment to better understand how environmental and visual stress markers—such as changes in color or growth—could reveal plant needs.

**Tracking Anyplant**

The Tracking Model for Anyplant builds upon the Tracking Anything Model(Yang et al. 2023), which integrates the Segment Anything Model (SAM) (Kirillov et al.2023) and XMem(Cheng et al. 2022). Figure 2 illustrates the model's structure. SAM generates the initial mask for the target object, and XMem performs semi-supervised video object segmentation to track the target across video frames. The Tracking Anything Model includes a user-friendly interface developed with Gradio(Abid et al. 2019), facilitating ease of use.

The Tracking Anything Model was extended with plant-specific functionalities. First, after XMem completes the semi-supervised video object segmentation of the plant, the average values of the Red (R), Green (G), and Blue (B) channels are calculated for each mask corresponding to non-overlapping tracked objects. Additionally, the area, width, and height of each mask for these objects are computed. Note that the calculated area, width, and height represent relative values due to the use of 2D images. The results are stored in a CSV file, and the file path is displayed in the user interface. Second, the G-channel values for each pixel within the mask of the tracked target object are calculated. Mapping these values to a predefined 256-color gradient generates the corresponding colors. These colors are superimposed onto the target objects in the original frames and blended into heatmap frames with transparency. This process is repeated for all video frames containing the target object, resulting in a heatmap video, which is displayed in the user interface. The color bar is also saved. Figure 3 presents the user interface.

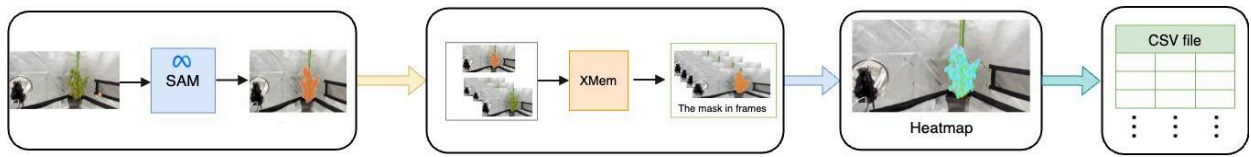

Figure 2. The framework of Tracking Anyplant

**Image Processing and Phenotypic Data Extraction** The collected images were processed using advanced computer vision techniques with the Segment Anything Model (SAM) for precise segmentation and the Track Anything Model (TAM) for tracking temporal changes. SAM enabled efficient mask extraction with minimal manual input, while TAM monitored changes like leaf growth and structural modifications. Key phenotypic features, including mean RGB values, mask areas, and plant dimensions, were extracted to assess plant health and growth. The data were systematically stored in CSV files for further analysis and integration.

**Data Integration** The environmental data collected from the IAQ Omni sensor and the phenotypic features extracted from the images were synchronized and merged into a unified dataset. This combined CSV file served as the foundation for correlating plant health indicators with environmental conditions and for predictive analysis.

**Predictive Modeling Using Lag-Llama foundation model** Lag-Llama, a state-of-the-art time-series forecasting model, was selected for its exceptional ability to handle irregular and noisy data, which are common challenges in environmental monitoring systems like those used in controlled plant growth experiments. Its design leverages advanced latent variable models to integrate contextual information across varying timescales, enabling effective multi-horizon forecasting while efficiently capturing both short-term fluctuations and long-term trends. Unlike models such as TFT, which rely on attention mechanisms for temporal dependencies and interpretability, Lag-Llama's architecture excels in encoding temporal dynamics and adapting to multivariate inputs, including environmental sensor readings (e.g., temperature, humidity, VOC levels) and image-derived phenotypic data (e.g., RGB, mask area). This comprehensive approach ensures accurate predictions of plant health metrics over extended periods while maintaining computational efficiency, making it ideal for applications in resource-constrained environments.

**The Architecture of Lag-Llama** The Lag-Llama builds upon Llama (Rasul et al., 2024), an open-source large language model. Figure 4 illustrates the architecture of Lag-Llama. The model processes historical time steps $x^i_{t-l_1}$, $x^i_{t-l_2}$, ..., $x^i_{t-l_C}$ from the univariate time series xi along with their corresponding covariates $c^i_{t-l_1}$, $c^i_{t-l_2}$, ..., $c^i_{t-l_C}$. The lag indices $\mathcal{L} = \{l_1, l_2, ..., l_C\}$ specify the time steps used for lagged features, where C represents the number of lagged features. The Projection layer maps the inputs into a high-dimensional space, transforming them into embedded features suitable for the Transformer's input format. Positional Encoding[] captures the sequential relationships between time steps, helping the model recognize the temporal structure of the data. The embedded features are processed through M Masked Transformer Decoder layers, where each layer employs a self-attention mechanism to model dependencies among features. Masking ensures the model considers only the current and previous time steps. The final output is the predicted distribution $P(x^i_{t+1})$ for the next time step $x^i_{t+1}$, generated via a distribution header that uses the Student-t distribution to quantify prediction uncertainty. The performance of the Lag-Llama model is primarily evaluated using the Continuous Ranked Probability Score (CRPS) (Matheson and Winkler 1976), a widely recognized metric in probabilistic forecasting literature.

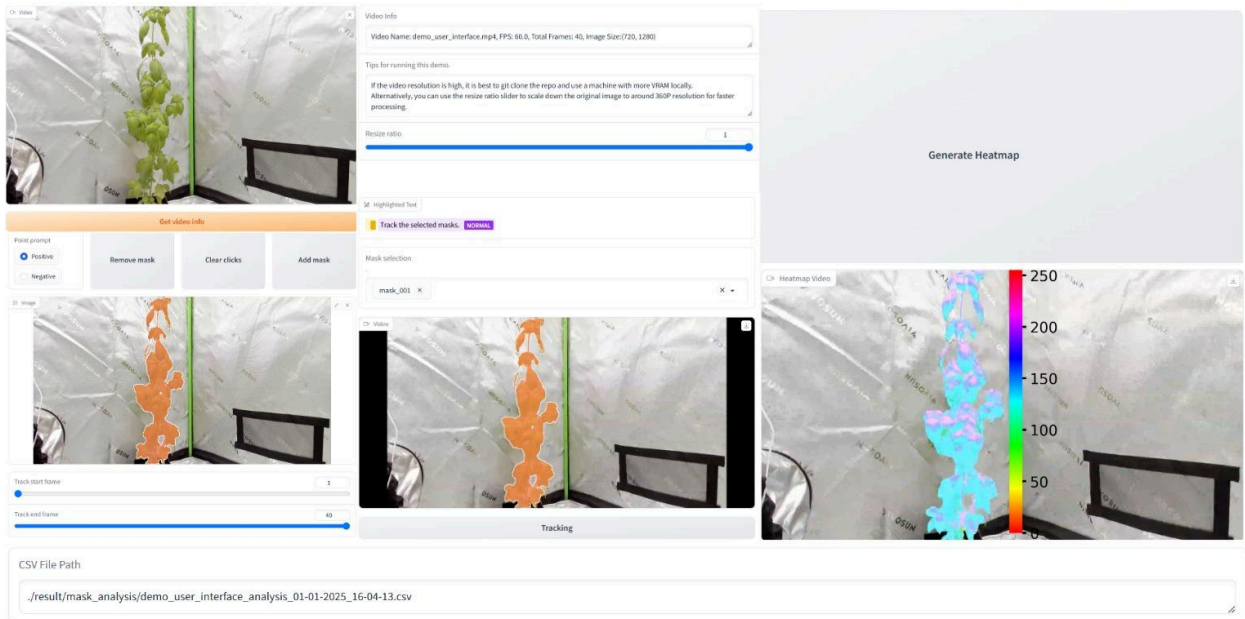

**Figure 3. User Interface Framework for the AnyPlant Tracking Model: A video is fed into the model, then adding a mask on the whole plant and tracking it to get the data.csv**

**Data Processing and Input Representation** The input data, comprising environmental parameters such as temperature, humidity, light intensity, and VOC, were preprocessed to address missing values and normalize the range. The preprocessing pipeline included:
- Imputation of missing values using a spline-based interpolation method.
- Replace Mask Area, Mask Height, and Mask Width with Area to Height Ratio and Height to Width Ratio

Lag-Llama processes input features in sequential batches, using overlapping fixed-length windows to preserve temporal structure and effectively capture short- and long-term dependencies.

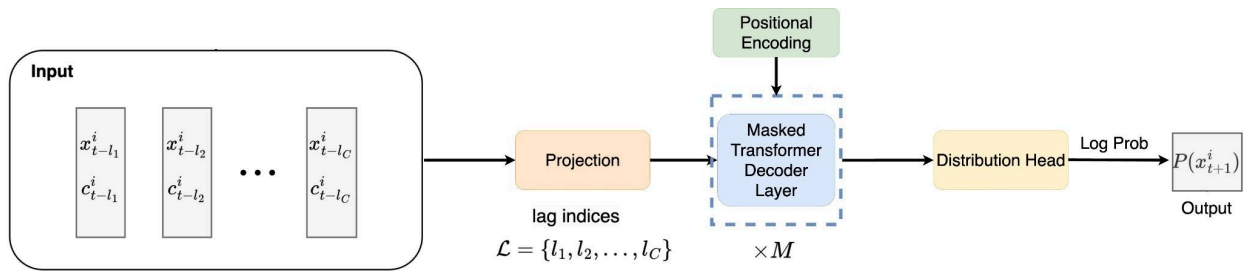

**Figure 4. Structure of Lag-Llama**

**Training Procedure** The Lag-Llama model was trained using supervised learning with Fine-tuning strategies to enhance forecasting accuracy for plant health metrics. Mean Squared Error (MSE) served as the loss function for continuous numerical predictions, while the Adam optimizer, with an initial learning rate of 0.001 and a cosine annealing schedule, dynamically adjusted learning rates. To prevent overfitting, dropout layers, L2 regularization, and early stopping (triggered after 20 epochs without validation loss improvement) were implemented.

**Evaluation and Validation** The performance of the Lag-Llama model was evaluated using Root Mean Square Error (RMSE) and correlation analyses to quantify deviations between predicted and observed values, focusing on the impact of environmental stressors, such as VOC levels, on plant phenotypic features like RGB changes and leaf area. Segmentation accuracy was validated by comparing automated outputs against manually annotated ground truth masks, ensuring the precision and robustness of the system.

**Sustainability and Resiliency Integration** The system demonstrated its ability to detect and predict plant stress through randomized watering schedules, enabling early interventions and reducing unnecessary water consumption. Integrating this framework into Building Management Systems (BMS) supports energy-efficient indoor gardening, promoting sustainable and resilient urban agriculture.

**Code and Data Availability** To encourage transparency and reproducibility, the code for the tracking-Anyplant along with the synchronized dataset of environmental and visual data, is publicly available on [GitHub](#) for future research and collaboration.

**RESULT and DISCUSSION**

**Result**

Experiments were conducted using Lag-Llama's Zero-shot and Fine-tuning approaches across various data combinations: RGB data, RGB with area-to-height and height-to-width ratios, RGB with environmental data, and RGB with both ratios and environmental data. Zero-shot predicts plant health metrics using pre-trained knowledge without requiring additional training on specific datasets. In contrast, Fine-tuning trains the pre-trained Lag-Llama model using multimodal datasets. Table 1 illustrates performance metrics for different data combinations under Zero-shot and Fine-tuning scenarios. Figure 5 visualizes the predictive performance of the Lag-Llama time-series model for both approaches. In the Zero-shot approach (subplots a–d), model performance improves as additional data modalities are incorporated (CPRS: 0.00145 to 0.00109; MSE: 0.024078 to 0.009680; MAE: 0.144201 to 0.096282). The combination of RGB, ratios, and environmental data achieves the best results (CPRS = 0.00109, MSE = 0.009680, MAE = 0.096282). The Fine-tuning approach (subplots e–h) exhibits a similar trend, with performance improving as data modalities increase (CPRS: 0.02219 to 0.00742; MSE: 4.780964 to 0.420777; MAE: 2.08486 to 0.595428). The integration of all available data yields the best results (CPRS = 0.00742, MSE = 0.420777, MAE = 0.595428). However, Fine-tuning underperforms compared to Zero-shot. For the best performance metrics, Zero-shot surpasses Fine-tuning (CPRS: 0.00109 vs. 0.00742; MSE: 0.009680 vs. 0.420777; MAE: 0.096282 vs. 0.595428).

Table 1. Performance Metrics Across Zero-Shot(Predictions without additional training on new data) and Fine-Tuning(further training on specific dataset) Scenarios

| Model | CPRS | MSE | MAE |
|---|---|---|---|
| Zero-shot-RGB | 0.00145 | 0.024078 | 0.144201 |
| Zero-shot-RGB with ratios | 0.00129 | 0.015242 | 0.125111 |
| Zero-shot-RGB with environmental data | 0.00124 | 0.010245 | 0.118096 |
| Zero-shot-RGB with ratios and environmental data | **0.00109** | **0.009680** | **0.096282** |
| Fine-tuning-RGB | 0.02219 | 4.780964 | 2.08486 |
| Fine-tuning-RGB with ratios | 0.01183 | 1.363109 | 1.18545 |
| Fine-tuning-RGB with environmental data | 0.01507 | 2.537505 | 1.498400 |
| Fine-tuning-RGB with ratios and environmental data | **0.00742** | **0.420777** | **0.595428** |

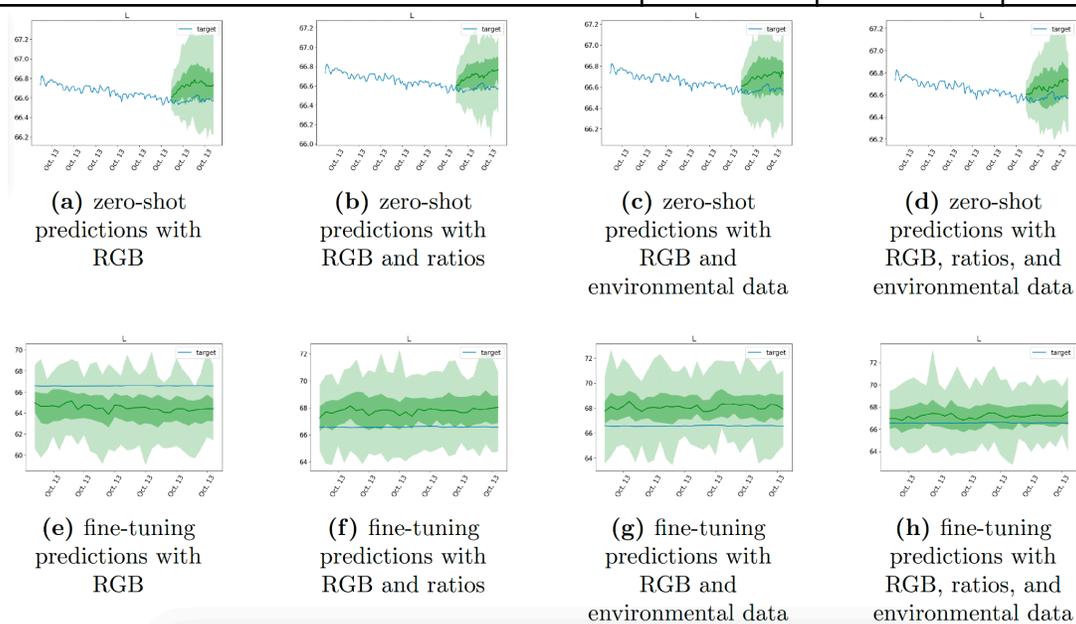

(a) zero-shot predictions with RGB

(b) zero-shot predictions with RGB and ratios

(c) zero-shot predictions with RGB and environmental data

(d) zero-shot predictions with RGB, ratios, and environmental data

(e) fine-tuning predictions with RGB

(f) fine-tuning predictions with RGB and ratios

(g) fine-tuning predictions with RGB and environmental data

(h) fine-tuning predictions with RGB, ratios, and environmental data

Figure 5. Visualization of Forecast Performance Across Feature Sets

**Discussion**

A detailed analysis of Zero-shot and Fine-tuning approaches across various data combinations reveals that multimodal data integration achieves the best results for both methods. These findings highlight the effectiveness of combining diverse data sources from Tracking AnyPlant and the Vivosun Smart Grow Tent Kit. Multimodal data integration captures the complex dynamics of water stress, emphasizing the synergy between visual, physiological, and environmental data.

The Tracking AnyPlant model, combined with the environmental data provided by the Vivosun Smart Grow Tent Kit, offers a powerful and effective solution for plant research. The integration of these tools demonstrates the ability to study plant growth, health, and stress responses under controlled conditions.

Zero-shot achieves superior performance and efficiency by eliminating the need for additional training, validating it as a reliable method for predicting plant water stress through multimodal data integration. The lower performance of Fine-tuning likely results from the dataset's temporal granularity. The training dataset, recorded at minute intervals, may not align well with the pre-trained Lag-Llama datasets. Pre-training the time-series foundation model on datasets with minute-level granularity could address this issue and potentially enable Fine-tuning to outperform Zero-shot. The reduced uncertainty and improved alignment with target values in the multimodal Zero-shot configuration underscore its potential for real-world applications, including plant health monitoring and irrigation optimization.

The proposed system integrates Tracking AnyPlant, the Vivosun Smart Grow Tent Kit, and the Lag-Llama foundation model to effectively predict plant water stress. This framework offers a reliable method for future plant research, such as growth and health prediction.

**CONCLUSION**

This study introduces an innovative framework for plant health monitoring that integrates computer vision, environmental sensors, and the Lag-Llama time-series foundation model. By analyzing multimodal data—including RGB imagery, physiological ratios, and environmental conditions such as temperature and humidity—the framework effectively predicts plant water stress and related metrics in controlled environments. The practical implications of this approach are significant for indoor and controlled agricultural settings. The system enables precise, real-time monitoring of plant health, allowing for timely interventions to optimize irrigation schedules and reduce resource consumption. The integration of visual, physiological, and environmental data into a cohesive framework paves the way for smarter and more efficient plant care systems, potentially enhancing crop yields and sustainability in urban farming and vertical agriculture. The ability of the framework to deliver high accuracy with zero-shot predictions underscores its scalability and applicability in diverse settings without the need for extensive retraining. This feature is particularly valuable for deploying the system in resource-constrained environments, where computational efficiency and adaptability are critical. Future steps for this research include expanding the dataset to encompass a wider variety of plant species and environmental conditions, refining the model to address temporal granularity mismatches in Fine-tuning scenarios, and integrating additional sensors for more comprehensive data collection. Additionally, incorporating this system into automated building management systems can further enhance sustainable practices, contributing to resilient, energy-efficient urban agriculture. By bridging gaps in existing methodologies, this framework establishes a foundation for advancing AI-driven plant health monitoring and aligns with global efforts toward sustainable and resilient food systems.